\PassOptionsToPackage{unicode}{hyperref}
\PassOptionsToPackage{hyphens}{url}
\documentclass{article}

\usepackage{amsmath,amssymb}
\usepackage{xcolor}

\usepackage{iftex}
\ifPDFTeX
  \usepackage[T1]{fontenc}
  \usepackage[utf8]{inputenc}
  \usepackage{textcomp}
  \usepackage{float}
  \usepackage{lmodern} 
\else
  \usepackage{unicode-math} 
  \defaultfontfeatures{Scale=MatchLowercase}
  \defaultfontfeatures[\rmfamily]{Ligatures=TeX,Scale=1}
\fi

\IfFileExists{microtype.sty}{\usepackage[]{microtype}
  \UseMicrotypeSet[protrusion]{basicmath}}{}

\makeatletter
\@ifundefined{KOMAClassName}{
  \IfFileExists{parskip.sty}{\usepackage{parskip}}{
    \setlength{\parindent}{0pt}
    \setlength{\parskip}{6pt plus 2pt minus 1pt}}
}{\KOMAoptions{parskip=half}}
\makeatother

\usepackage{longtable,booktabs,array,calc}
\usepackage{etoolbox}
\makeatletter
\patchcmd\longtable{\par}{\if@noskipsec\mbox{}\fi\par}{}{}
\makeatother
\IfFileExists{footnotehyper.sty}{\usepackage{footnotehyper}}{\usepackage{footnote}}
\makesavenoteenv{longtable}
\usepackage{graphicx}
\makeatletter
\newsavebox\pandoc@box
\newcommand*\pandocbounded[1]{%
  \sbox\pandoc@box{#1}%
  \Gscale@div\@tempa{\textheight}{\dimexpr\ht\pandoc@box+\dp\pandoc@box\relax}%
  \Gscale@div\@tempb{\linewidth}{\wd\pandoc@box}%
  \ifdim\@tempb\p@<\@tempa\p@\let\@tempa\@tempb\fi
  \ifdim\@tempa\p@<\p@\scalebox{\@tempa}{\usebox\pandoc@box}%
  \else\usebox{\pandoc@box}\fi}
 \def\fps@figure{htbp}
\makeatother

\ifPDFTeX
  \newcommand{\RL}[1]{\beginR #1\endR}
  \newcommand{\LR}[1]{\beginL #1\endL}
  \newenvironment{RTL}{\beginR}{\endR}
  \newenvironment{LTR}{\beginL}{\endL}
\fi
\ifluatex
  \newcommand{\RL}[1]{\bgroup\textdir TRT#1\egroup}
  \newcommand{\LR}[1]{\bgroup\textdir TLT#1\egroup}

\fi

\IfFileExists{xurl.sty}{\usepackage{xurl}}{}
\usepackage{hyperref}
\hypersetup{hidelinks,pdfcreator={LaTeX via pandoc}}
\usepackage{bookmark}

\usepackage{authblk}
\makeatletter
\renewcommand\AB@affilsepx{ \par }  
\makeatother

\title{\textbf{Ensembling Multilingual Transformers for Robust Sentiment Analysis of Tweets } }
\author[a,b]{Meysam Shirdel Bilehsavar}
\author[c]{Negin Mahmoudi}
\author[d]{Mohammad Jalili Torkamani}
\author[e]{Kiana Kiashemshaki}

\affil[a]{Department of Computer Science, University of South Carolina, USA}
\affil[b]{Artificial Intelligence Institute, University of South Carolina, USA}
\affil[c]{Department of Civil, Environmental, and Ocean Engineering, Stevens Institute of Technology, New Jersey, USA}
\affil[d]{School of Computing, University of Nebraska--Lincoln, Lincoln, Nebraska, USA}
\affil[e]{Department of Computer Science, Bowling Green State University, Bowling Green, Ohio, USA}

\affil[]{\textit{Corresponding author:} \href{mailto:Meysam@email.sc.edu}{Meysam@email.sc.edu}}
\affil[]{\textit{Contributing authors’ emails: },
\href{mailto:Meysam@email.sc.edu}{Meysam@email.sc.edu}
\href{mailto:nmahmoud1@stevens.edu}{nmahmoud1@stevens.edu},
\href{mailto:mJaliliTorkamani2@huskers.unl.edu}{mJaliliTorkamani2@huskers.unl.edu},
\href{mailto:kkiana@bgsu.edu}{kkiana@bgsu.edu}
}

\date{} 

\begin{document}
\maketitle

\section{Abstract}\label{abstract}

Sentiment analysis is a very important natural language processing activity in which one identifies the polarity of a text, whether it conveys positive, negative, or neutral sentiment. Along with the growth of social media and the Internet, the significance of sentiment analysis has grown across numerous industries such as marketing, politics, and customer service. Sentiment analysis is flawed, however, when applied to foreign languages, particularly when there is no labelled data to train models upon. In this study, we present a transformer ensemble model and a large language model (LLM) that employs sentiment analysis of other languages. We used multi languages dataset. Sentiment was then assessed for sentences using an ensemble of pre-trained sentiment analysis models: bert-base-multilingual-uncased-sentiment, and XLM-R. Our experimental results indicated that sentiment analysis performance was more than 86\% using the proposed method.

Keywords: Sentiment Analysis; Ensemble Language Models; Multilingual BERT; XLM-R

\section{Introduction}\label{introduction}

Emotional expression is subjective, shaped by cultural background, emotional intelligence, individual personality traits and characteristics. While we can communicate emotions clearly in verbal communication due to nuances of tone and inflection, it's harder to determine how a person feels and what emotions they're expressing through words alone. the rapid growth of social media enables us to analyze user sentiment \cite{cui2023survey}. Despite extensive work on social-media sentiment analysis, multilingual coverage remains limited for mixed datasets that include low-resource languages (e.g., Arabic) because annotated data, reliable tools, and standardized benchmarks are scarce.\cite{kumar2021sentiment}.

Over the past decade, Natural Language Processing (NLP) has grown rapidly, and sentiment analysis has emerged as one of its key areas. This technique is now widely applied in fields such as market intelligence, customer feedback analysis, social-media monitoring \cite{dezhboro2025analyzing}, and reputation management.\cite{manias2023multilingual}. The capability to extract sentiments from
textual information has become invaluable for organizations and
scientists aiming to derive useful insights from the large pool of user-generated data. This project explores sentiment analysis with a
specific focus on the intricate context of multilingual tweet texts
\cite{wadhawan2021arabert}. The objective is to classify the sentiments of these brief and dynamic tweets into three categories: positive, negative, or neutral. Sentiment analysis enables a broad range of applications, providing insights that support decision making and trend analysis in areas such as business and finance\cite{ataei2025systematic}.

Large Language Models (LLMs) have emerged as foundational architectures within the field of NLP, driving state-of-the-art performance across a range of tasks, including text classification \cite{wei2023empirical, chae2023large}, conversational agents \cite{demetriadis2023conversational, liao2023proactive, abbasian2024conversationalhealthagentspersonalized}, and sentiment analysis \cite{miah2024multimodal}. Their capacity to model contextual semantics and facilitate cross-lingual transfer renders them particularly valuable for multilingual applications. Nonetheless, LLMs are susceptible to inheriting and, in some cases, amplifying the biases embedded in their training corpora, thereby raising concerns regarding fairness, reliability, and trustworthiness in practical deployments. Recent research works have proposed simulation-based approaches to mitigating such biases, offering strategies that enhance both the robustness and equity of multilingual sentiment analysis systems \cite{kiashemshaki2025simulating, zhou2024unibias}.

Despite the rapid growth of sentiment analysis research, an important challenge remains: how can effective methods be transferred to non-English languages where labeled data are scarce? For effective training, high-quality and diverse datasets are essential, as label noise can further degrade model performance \cite{mahjourian2025sanitizingmanufacturingdatasetlabels}.
This study addresses that challenge by proposing a comprehensive methodology and presenting empirical evidence that demonstrates the practicality and accuracy of cross-lingual sentiment analysis using
translation. In today\textquotesingle s age of high-frequency
social media usage, sentiment analysis is now a part and parcel,
particularly on microblogging websites such as Twitter. The amount of
user-generated content provides a rich source of understanding the
public opinions, comments, and sentiments \cite{gandhi2023multimodal}. Another prevalent
approach applied in sentiment analysis is applying machine learning
classification techniques, as indicated in studies like the use of
distant supervision and noisy labels through emoticons and acronyms in
tweets \cite{das2023multimodal}. Applying this technique is one approach to balancing the
brevity and simplicity that feature tweets towards improved sentiment
analysis.

Linguistic diversity, cultural nuances, and geographically localized
patterns on social media sites complicate sentiment analysis.
Cross-lingual and multilingual approaches have emerged to overcome this
challenge, embodied by the work using XLM-RoBERTa for cross-lingual
sentiment analysis, English-to-Hindi knowledge transfer from a
resource-rich English to a resource-poor Hindi \cite{liao2021improved}. Such
adaptability is crucial in good sentiment analysis in different
linguistic contexts. Sentiment analysis is not limited to English-based
platforms either. Sentiment analysis studies for Arabic tweets note
efforts at combining pre-processing methods with transformer models like
AraELECTRA and AraBERT to address the subtleties of sarcasm and
sentiments in Arabic \cite{mercha2023machine}.

The article aims to provides an overview of Multilingual Sentiment
Analysis including definitions and methods. The remainder of the article
are as follows. Section 3 presents the proposed simulation framework for
Sentiment Analysis, outlining the experimental design, implementation
details, and evaluation metrics. Section 4 reports and analyzes the
results of the simulations, highlighting their implications model
performance. Finally, Section 5 and 6 concludes the paper by summarizing
key findings, discussing limitations, and suggesting directions for
future research.

\section{Concepts and related works}\label{concepts-and-related-works}

Sentiment analysis, more widely referred to as opinion mining, is one of
the most significant subfields of natural NLP in
recent decades. Sentiment analysis essentially determines if a text
carries positive, negative, or neutral sentiment \cite{oueslati2020review}.This analytical capability is today a necessity for gauging public opinion,
responding to customer feedback, and comprehending social conversation,
which in turn informs decision-making in domains as varied as marketing,
politics, and public administration. With digital communication networks
having spread exponentially, large volumes of text data are generated
every day, which further increases the need for effective automated
sentiment classification systems \cite{chan2023state}.

While English has been the biggest recipient of huge labeled datasets
and heavily tuned language models, enabling researchers to have an
extremely large building block to work with, a huge gap in resources
and performance persists for other languages. The explosion in usage of
social media websites, particularly over websites like Twitter, has
offered enormous reservoirs of user-generated text \cite{wahidur2024enhancing}. This
plethora of languages and dialects makes monolingual solutions
insufficient, summoning the dire need for sentiment analysis systems
that generalize well across languages with different scripts, syntactic
structures, and cultural sensitivities. Attempts at multilingual
solutions habitually encounter the problem of limited labeled data,
especially for low-resource languages \cite{xing2025designing}. Advances such as
BERT-based multilingual classifiers and zero-shot models have
demonstrated encouraging adaptability, leveraging cross-lingual transfer
via shared embedding spaces. Experiments have also moved towards
strengthening these systems by trying various pretraining corpora,
multilingual-specific hyperparameter tuning, and optimizing data
harvesting procedures for better model generalization \cite{xu2024reasoning}.

Machine translation (MT) technologies have played an important role in
extending the capabilities of sentiment analysis beyond high-resource
languages \cite{uddin2024explainable}. Within information retrieval, researchers such as
Wankhade et al have investigated the compromises of using translated
queries and reported a moderate performance reduction relative to
monolingual retrieval, approximately 10--15\% relative to the language
pair, but sometimes weaker \cite{wankhade2022survey}. Cross-lingual summarization
research has pioneered the integration of MT with content ranking, such
as Wan et\,al.\textquotesingle s ranking of source-language sentences
before translation to ensure content quality \cite{cao2020jointly}, and wang
et\,al.\textquotesingle s inversion of this sequence to rank
translations first. Both demonstrate the applicability of MT as a
bridging mechanism, yet with the constraint of translation noise
\cite{wang2022survey}.

Current sentiment analysis methods can be broadly divided into three
categories: corpus-based, lexicon-based, and hybrid approaches.
Corpus-based methods primarily rely on labeled corpora to train machine
learning classifiers, ranging from traditional SVM and Naïve Bayes
models to newer transformer-based architectures \cite{catelli2022lexicon}. Lexicon-based
methods leverage manually crafted lists of sentiment-carrying
words, perhaps augmented with part-of-speech tags or polarity
scores, and optionally combine them with syntactic rules \cite{qi2023sentiment}.
Hybrid approaches aim to combine the precision of lexicons with the
adaptability of machine learning, sometimes incorporating unlabeled
corpora to enhance coverage \cite{kaur2023deep}. The strengths and weaknesses of
each type are often a function of the availability and quality of
linguistic resources for the target language \cite{alantari2022empirical}.

Multilingual sentiment analysis must contend with the fact that social
media language is informal, rife with slang, nonstandard spellings, and
creative punctuation \cite{liu2024application}. Preprocessing is more than a preparatory
step; it is a performance determining factor for models. Common
operations include tokenization (splitting text into words or symbols)
\cite{nazir2025leveraging}, sentence splitting (identifying clause boundaries) \cite{mabokela2022multilingual},
stop-word removal (removal of highly frequent but semantically lean
words) \cite{garg2022text}, stemming (reducing words to root forms) \cite{das2023sentiment}, and
part-of-speech tagging (marking grammatical categories) \cite{chan2024exploring}. In
multilingual datasets, language idiosyncrasies, e.g., German compound
word formation or Arabic script variation, demand language-specific
preprocessing rules. The multilingual Twitter lexicon constructed by
Posadas-Durán et\,al. for English, Spanish, Dutch, and Italian shows the
usefulness of domain-specific tools in smoothing out such idiosyncrasies
\cite{manias2023multilingual}.

Several highly regarded sentiment lexicons underlie state-of-the-art
analysis pipelines. SenticNet \cite{ani2024multilingual}, is a concept-level polarity
detection method that uses common-sense reasoning for deeper semantic
understanding and offers multilingual adaptations. SentiWordNet
\cite{kanfoud2022senticode}, based on WordNet synsets, offers polarity scores ranging from
0.0 to 1.0, though it is poor when handling multi-word expressions or
idiomatic phrases.

In addition to lexicons, hand-annotated sentiment corpora \cite{nikolova2024evaluating}
provide training data for corpus-based methods. Multimodal corpora like
the YouTube corpus, coupling audio, video, and transcript data, present
additional complexity in the guise of ambient noise and speaker
diversity. The MPQA Subjectivity Lexicon offers wide coverage of
subjective words with part-of-speech tags and polarity intensities.
Availability of these corpora allows us to construct models with
expertise in specific genres, domains, or modalities, a consideration
crucial for multilingual adaptation \cite{kanfoud2021linking}. Table 1 shows the overview
of Cross-Lingual and Multilingual Sentiment Analysis Techniques.

\begin{longtable}[]{@{}
  >{\centering\arraybackslash}p{(\linewidth - 6\tabcolsep) * \real{0.2039}}
  >{\centering\arraybackslash}p{(\linewidth - 6\tabcolsep) * \real{0.2418}}
  >{\centering\arraybackslash}p{(\linewidth - 6\tabcolsep) * \real{0.2962}}
  >{\centering\arraybackslash}p{(\linewidth - 6\tabcolsep) * \real{0.2581}}@{}}
\caption{Overview of Cross-Lingual and Multilingual
Sentiment Analysis Techniques}\tabularnewline
\toprule\noalign{}
\begin{minipage}[b]{\linewidth}\centering
\textbf{Language(s)}
\end{minipage} & \begin{minipage}[b]{\linewidth}\centering
\textbf{Approach Type}
\end{minipage} & \begin{minipage}[b]{\linewidth}\centering
\textbf{Key Techniques / Features}
\end{minipage} & \begin{minipage}[b]{\linewidth}\centering
\textbf{Notes}
\end{minipage} \\
\midrule\noalign{}
\endfirsthead
\toprule\noalign{}
\begin{minipage}[b]{\linewidth}\centering
\textbf{Language(s)}
\end{minipage} & \begin{minipage}[b]{\linewidth}\centering
\textbf{Approach Type}
\end{minipage} & \begin{minipage}[b]{\linewidth}\centering
\textbf{Key Techniques / Features}
\end{minipage} & \begin{minipage}[b]{\linewidth}\centering
\textbf{Notes}
\end{minipage} \\
\midrule\noalign{}
\endhead
\bottomrule\noalign{}
\endlastfoot
Multiple & Cross-lingual IR & MT for query translation, performance
comparison & \textasciitilde10--15\% drop vs monolingual \\
Multiple & Cross-lingual summarization & Ranking + MT quality scoring +
informativeness measure, sentence selection pre/post translation &
Balances quality \& informativeness \\
Multiple & Cross-lingual summarization & Translate → rank → select &
Enhances readability \\
Czech & Corpus-based & POS tagging, unigrams/bigrams, emoticons, Delta
TF-IDF, SVM, MaxEnt & F-measure = 0.69 \\
Chinese & Corpus-based & POS tagging, feature selection (DF, CHI, MI,
IG), SVM, NB, k-NN, centroid, winnow & SVM best, IG most
discriminative \\
French & Corpus-based & Translation to English, lemmatization, POS
tagging, SentiWordNet semantic orientation, SVM & \textasciitilde94\%
accuracy, translation loss issue \\
Multiple (IT, DE, ES, FR) & Hybrid (MT + ML) & Minimal preprocessing,
SVM SMO, n-grams/bigrams, MT dictionary creation & Highest accuracy in
English \\
Arabic & Corpus-based & Tokenization, stemming, bigrams, negation
handling, SVM, NB, k-NN & SVM best; small training set \\
EN, ES, NL, IT & Lexicon-based preprocessing & Multilingual lexicon for
Twitter \& social media normalization & Improves preprocessing
efficiency \\
Multiple & Lexicon-based & Concept-level polarity analysis, common-sense
reasoning & Multilingual versions available \\
Multiple & Lexicon-based & Synset polarity scoring & No phrase-level
polarity \\
Spanish & Lexicon-based & Affect dictionary with Probability Factor of
Affective use & Manual annotation by 19 annotators \\
\end{longtable}

\section{Methodology}\label{methodology}

The goal of this study is to develop and evaluate a multilingual sentiment analysis model able to classify tweeter comment texts as negative, neutral, or positive for a diverse set of languages. The general workflow involves a number of steps: dataset collection and annotation, preprocessing, sentiment label mapping, dataset split, multilingual tokenization, model training, and evaluation. By leveraging transformer-based multilingual models, the paper discusses the performance of state-of-the-art natural language processing architectures in crosslinguistic sentiment detection. 

Dataset for this study was obtained from the Kaggle open repository, under name of "multilingual sentiment analysis" authored by Suraj. The selected dataset consisted of multilingual tweets with sentiment ratings on a five-star scale as well as language identifiers. A sample of rows within the dataset is presented in Table 1.

\begin{longtable}[]{@{}
  >{\centering\arraybackslash}p{(\linewidth - 6\tabcolsep) * \real{0.10}}
  >{\arraybackslash}p{(\linewidth - 6\tabcolsep) * \real{0.55}}
  >{\centering\arraybackslash}p{(\linewidth - 6\tabcolsep) * \real{0.15}}
  >{\centering\arraybackslash}p{(\linewidth - 6\tabcolsep) * \real{0.20}}@{}}
\caption{Sample rows from the multilingual tweet dataset}\tabularnewline
\toprule\noalign{}
\begin{minipage}[b]{\linewidth}\centering
\textbf{Index}
\end{minipage} & \begin{minipage}[b]{\linewidth}\centering
\textbf{Tweet}
\end{minipage} & \begin{minipage}[b]{\linewidth}\centering
\textbf{Language}
\end{minipage} & \begin{minipage}[b]{\linewidth}\centering
\textbf{Sentiment}
\end{minipage} \\
\midrule\noalign{}
\endfirsthead
\toprule\noalign{}
\begin{minipage}[b]{\linewidth}\centering
\textbf{Index}
\end{minipage} & \begin{minipage}[b]{\linewidth}\centering
\textbf{Tweet}
\end{minipage} & \begin{minipage}[b]{\linewidth}\centering
\textbf{Language}
\end{minipage} & \begin{minipage}[b]{\linewidth}\centering
\textbf{Sentiment}
\end{minipage} \\
\midrule\noalign{}
\endhead
\bottomrule\noalign{}
\endlastfoot
0 & Lionel Messi, que ha estado vinculado con un traslado a Barcelona, es probable que se mude a Barcelona este verano, con un posible traslado & es & 3 stars \\
1 & This is a guest post by The Joy of Truth. To read more of her essays on the topic, check out her new book, The Joy & en & 4 stars \\
2 & Nous sommes tous conscients de la popularité d'Internet. Ceci est dû au fait qu'il s'agit d'un marché énorme, avec un & fr & 5 stars \\
3 & El baño en el sistema de metro de la ciudad de Nueva York también es una parte clave de este proyecto & es & 4 stars \\
\end{longtable}

The foundation of the analysis was a three column, multilingual table of tweets that included tweet, language, and sentiment. Each row contained raw tweet text, the language tag for that text, and an annotated sentiment rating on a 1 to 5-star scale, where 1 star was extremely negative sentiment and 5 stars was extremely positive sentiment, with the mid-ratings referring to nuanced levels of emotional polarity. The corpus was structured to be balanced across sentiment categories and topical and linguistic variation so that the corpus captured various communicative styles between languages without over-representation of strong labels. Such balance was considered essential to train stable and generalizable models. Because of the noisy nature of social media text, an end-to-end preprocessing pipeline was used. Unnecessaries such as hyperlinks, user tags, unusual punctuation, excessive whitespace, and non-alphabetic symbols were eliminated in a controlled way. In addition, normalization steps guaranteed greater uniformity across the dataset, which was particularly important because of the wide variety of languages and orthographic conventions involved. These steps preserved only semantically relevant content for downstream processing.

The original five-star sentiment scale was later collapsed into a three-class categorical format: negative for the ratings 1–2, neutral for rating 3, and positive for the ratings 4–5. This collapse yielded a better interpretable classification system and reduced sparsity in middle categories. Figure 1, "Overall Sentiment Distribution," illustrates percentage distribution of these three sentiment classes across the entire dataset. By aggregating all languages into a single high-level view, the figure presents a simple overview of emotional polarity in the data. The high-level view is something that can serve as a benchmark by which to compare more subtle, language-by-language analysis so that researchers can find global sentiment balance rather than variation on language-by-language terms.

\begin{figure}
\centering
\includegraphics[width=4.13277in,height=2.40336in]{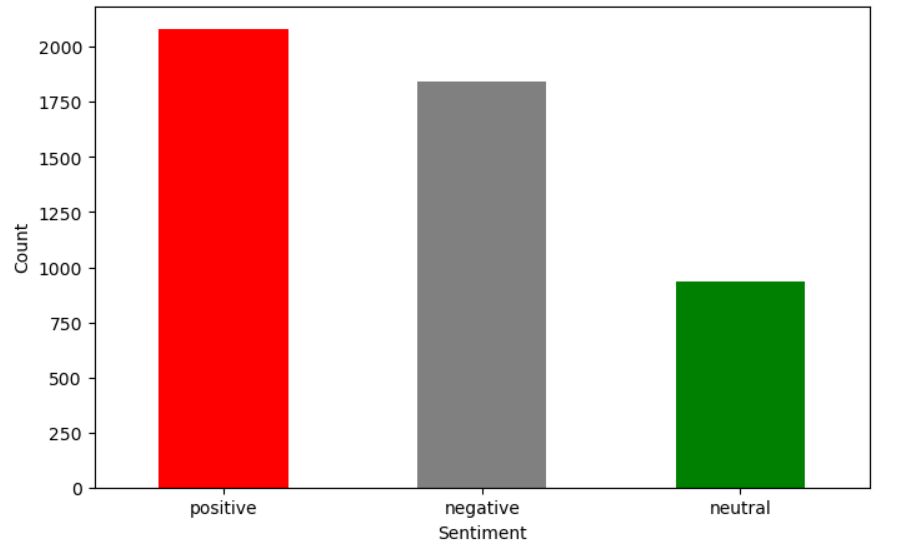}
\caption{Overall Sentiment Distribution}
\end{figure}

while Figure 1 shows the overall sentiment distribution, Figure 2, titled Sentiment Distribution by Language, is a disaggregated one where every bar shows the relative sizes of negative, neutral, and positive categories for one language subset. This visualization highlights sentiment balance differences potentially brought about by cultural tendencies, linguistic conventions, or topical tweet subject matters in each language. For example, there will be a greater percentage of neutral posts in some languages due to politeness standards or indirect communication modes, while others will convey more polarities. In comparing Figure 2 with the overall perspective of Figure 1, the study identifies where sentiment distribution corresponds or diverges from the overall dataset, providing more access to multilingual heterogeneity in emotion expression.
\begin{figure}
\centering
\includegraphics[width=4.29224in,height=2.8429in]{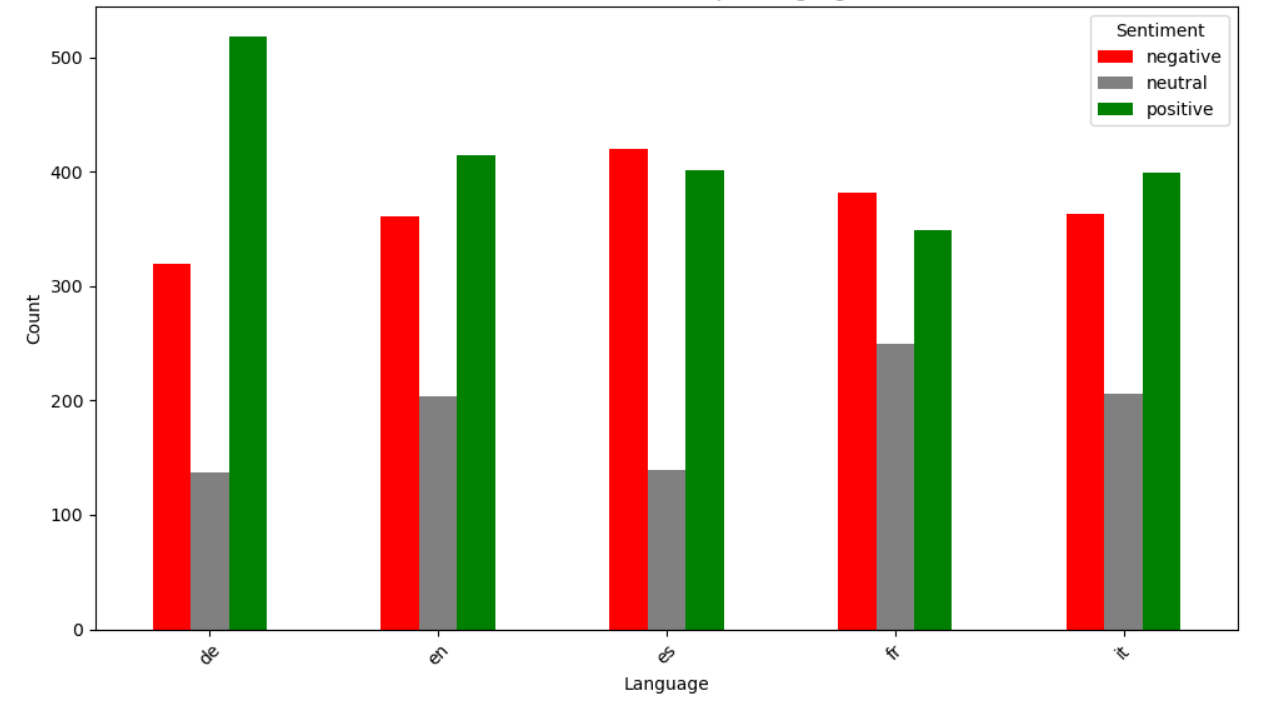}

\caption{Sentiment Distribution by Language}
\end{figure}

To ensure that the modeling process was statistically sound, the dataset was split into training, validation, and test subsets by stratified sampling. The process was done jointly across sentiment labels and language labels to keep proportional representation in each split. The use of stratification avoided imbalances that could otherwise affect model training or balloon evaluation scores, ensuring sentiment diversity and language diversity were kept consistently across subsets. Following the partitioning, tweets were tokenized with multilingual transformer-based models such as mBERT (Multilingual BERT) and XLM-R (XLM-RoBERTa). Tweets were transformed into input token ID sequences and attention masks for individual tweets. Sequences exceeding the max model length were truncated to enhance computational effectiveness without affecting semantic representation drastically. The classifier was built using the AutoModelForSequenceClassification architecture of the Hugging Face Transformers library. Training was performed using the AdamW optimizer, along with a linear learning rate scheduler to provide stable convergence. Model performance was tracked every epoch at the end of each epoch on the validation set, using accuracy, macro-averaged precision, recall, and F1-score as performance metrics.

The trained model was tested subsequently using the held-out test set. In addition to overall accuracy and macro metrics, a complete classification report and a confusion matrix were prepared to investigate patterns of misclassification. To quantify variation between language groups, per-language evaluations were also conducted and the results were charted in "Confusion Matrix per Language" charts. These plots not only revealed which classes were being most frequently misclassified, but also how their misclassifications varied across languages, providing a more detailed view of the model's relative strengths and weaknesses in multilingual environments. A visual overview of the entire pipeline is presented in Figure 3, which is titled "Proposed Workflow." The flowchart presents orderly stages of the study: from preprocessed raw data, through mapping sentiment labels, stratified dataset partitioning, tokenization across languages, model training and validation, and to end-to-end testing and per-language evaluation. This overview presents a clear-cut methodological map that encapsulates the broad scope of the study.

\begin{figure}
\centering
\includegraphics[width=3.89326in,height=8.20833in]{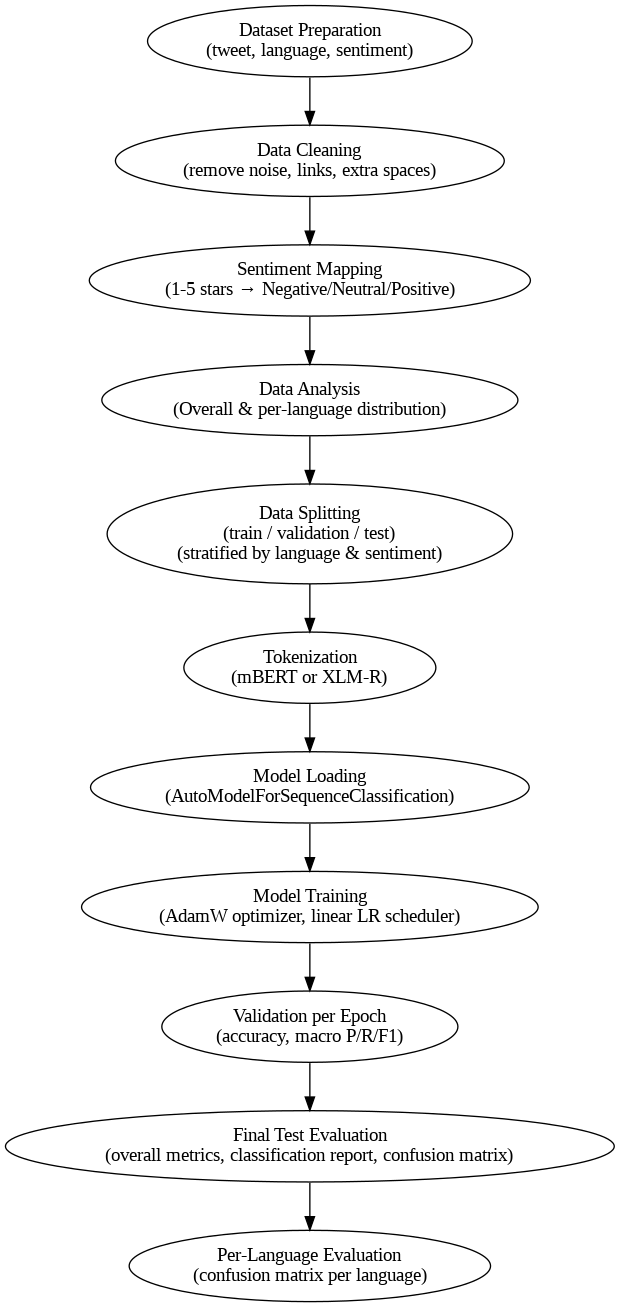}
\caption{Flowchart of proposed method}
\end{figure}

Despite this organized approach, there were some serious problems. First, the phenomenon of code-switching, by which users change languages in a tweet, presented challenges to tokenization and sentiment analysis. Second, the model had difficulties with sarcasm and irony, as these forms of speech are culturally nuanced and could not be modeled using only lexical signals. Third, the differences in performance between high-resource and low-resource languages were observed: while well-represented languages such as English or Spanish had rich pre-training corpora, under-represented languages were tougher due to limited training resources. Fourth, non-standard orthography, spelling variations, and geographic dialects decreased tokenization accuracy, hence affecting downstream sentiment classification. Lastly, cultural differences in feeling expression influenced relative rates of sentiment categories in that some language groups had a tendency for more positive or more neutral reporting.

\section{Results}\label{results}

Training was done over three complete epochs, with performance monitored
on a held-out validation set after every epoch in order to monitor
progress and identify the onset of overfitting.

Epoch\,1 was validated at 82.75\% accuracy, with a macro precision of
83.12\%, macro recall of 82.74\%, and a macro F1-score of 82.66\%. These
numbers established that the model was capable of attaining a decent
baseline estimation of the multilingual sentiment patterns from the
start. Some misclassifications, however, indicated that more subtle
differences between classes, most prominently between "negative" and
"neutral," whose linguistic forms are generally neutral, were not yet
optimized.

Epoch 2 saw a big jump in model performance, with accuracy at 89.10\%.
Macro precision and recall were both 89.49\% and 89.09\% respectively,
as was the macro F1-score, which was 89.09\%. This was one of sharp
rise, showing that the model had learned more discriminative features
from the multilingual data effectively, reducing overlap between
sentiment classes and raising confidence in borderline cases.

On Epoch\,3, the accuracy in validation had reached 89.53\%, macro
precision 89.66\%, macro recall 89.53\%, and macro F1-score 89.50\%.
Now, there was convergence of the model: performance gains were smaller,
in line with diminishing returns of additional epochs. Precision and
recall remained practically constant for all classes, meaning that now
the decision boundaries were more robust and less prone to bias towards
any individual class.

Upon testing with the final test set, the model achieved an accuracy
rate of 90.05\%. Macro precision, recall, and F1-score were also well
balanced at 90.05\%--90.09\%, with weighted metrics almost similar,
reflecting the robustness of the model on classes with different sample
sizes of the multilingual data.

Classification report gave a more nuanced perspective of performance by
sentiment category:

\begin{itemize}
\item
  Negative: precision 0.89, recall 0.91, F1 0.90 (support: 388)
\item
  Neutral: precision 0.89, recall 0.91, F1 0.90 (support: 389)
\item
  Positive: precision 0.92, recall 0.88, F1 0.90 (support: 389)
\end{itemize}

These results show that "positive" predictions were most accurate, that
the model was optimal at avoiding false positives in this class, and
"negative" and "neutral" picked up a bit more recall, so fewer actual
samples went unclassified in these categories. This tradeoff is
extremely beneficial in multilingual sentiment analysis, where there is
class imbalance or cultural interpretation variation which can skew
predictions.

\begin{figure}
\centering
\includegraphics[width=3.7157in,height=3.37788in]{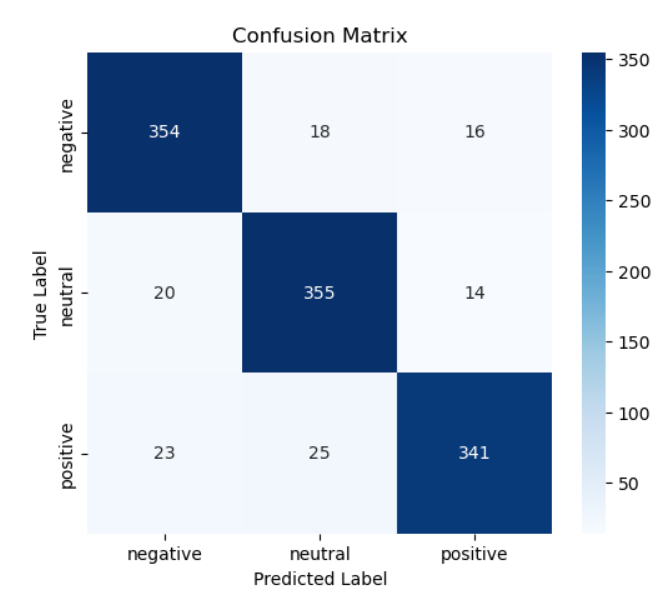}
\caption{Overall confusion matrix}
\end{figure}

Figure\,4 displays the whole confusion matrix, a visual tool to assist
with the identification of systematic misclassifications of patterns.
The prominent diagonal in the matrix accounts for correct predictions:
354 "negative," 355 "neutral," and 341 "positive" tweets correctly
classified. Misclassifications were low and fairly symmetrical:

\begin{itemize}
\item
  "Negative" tweets were misclassified as "neutral" (18 instances)
  or "positive" (16 instances).
\item
  "Neutral" tweets had 20 misclassifications as "negative" and 14 as
  "positive."
\item
  Positive" messages were mistakenly marked as "neutral" 25 times and as
  "negative" 23 times.
\end{itemize}

No systematic overprediction of any one class was seen, common pitfall
in multilingual data when all language-specific sentiment cues are
disparately present. This error pattern demonstrates that
misclassifications resulted less from structurally flawed
models\textquotesingle{} impaired acquired features and more from
intrinsically uncertain samples.

\section{Discussion}\label{discussion}

In this large multilingual sentiment analysis task, we achieved the
complexities of processing diverse tweet texts through a number of
transformer-based models, language-specific architectures, and ensemble
techniques. The results of the evaluation as represented by the
performance metrics identify a multi-dimensional trend of strengths and
weaknesses in both languages and approaches.

Model-level comparisons showed dramatic differences in performance.
RoBERTa was best on English sentiment analysis with good
precision, recall tradeoff and outperforming the vast majority of other
settings. In contrast, mBERT, designed for general multilingual
coverage, exhibited inconsistent results, as pointed out, multilingual
pretraining does not necessarily result in consistent accuracy over all
languages. AraBERTv02 achieved competitive accuracy on Arabic, though
the morphological richness and script nature of the language were
problematic for fine-grained sentiment discrimination.

Aside from single-model analysis, ensemble approaches were a key factor
in improving performance. The Majority Voting Ensemble and the newly
presented multi‑BERT ensemble both improved macro‑average F1-scores by
correcting for biases in a single model. This validates the application
of bagging-style aggregation to reduce variance and increase robustness,
especially in settings with heterogeneous linguistic inputs and sparse
per-language data. Furthermore, language‑independent
experiments, trained on combined sets such as
English\,+\,Arabic, predict promising generalization potential, albeit
at the cost of periodic language‑specific accuracy dips.

One frequent choice in experiments was the use of macro‑average F1 as
the main measure. Because class imbalance is present in sentiment data,
the measure ensured that evaluation centered on equal consideration of
all sentiment classes and thus prevented majority‑class dominance from
hiding minority class weakness. It was particularly helpful in the
multilingual setting, in which minority classes differ by language.

But the experiments also exposed the essential shortcomings of employing
machine-translated training data. Translations of poor quality increased
feature sparsity and rendered positive and negative examples less
discernible, with resulting losses in performance of up to eight
percentage points in the most adverse case. The effect was more
pronounced with methods such as AdaBoost, which were discovered
extremely sensitive to noisy inputs. Conversely, Bagging consistently
improved F-scores through decreased variance, predominantly for German
data with lowest translation quality and highest variance.

On a language-specific level, translation quality directly affected
learnability. For Spanish, where translation Bleu scores were
highest, the classifiers retained discriminative power and yielded
balanced class prediction. For lower-quality translations, classifiers
would reduce their predictions into a single sentiment class and even
lost the ability to differentiate between positive and negative
instances. French models declined severely but still managed to get some
degree of separation, while other low‑quality examples, such as some
German and Arabic translations, had severe degradation. Notably,
aggregating all translated datasets together added increased levels of
noise that impaired classification accuracy by overburdening training
with contaminated samples.

These findings draw attention to a double message: on the one hand,
high‑quality language resources remain fundamental to sentiment analysis
because even state‑of‑the‑art models cannot fully escape systematic
translation flaws; on the other hand, ensemble and variance‑reduction
techniques such as Bagging can limit, though not eliminate, these
artifacts. Together, the results validate that multilingual sentiment
analysis benefits from an effectively tuned trade‑off between
language‑specific specialisation, robust cross‑lingual design, and
prudent use of machine translation.

\section{Conclusion}\label{conclusion}

This research demonstrated that multilingual sentiment classification
for noisy real-world data such as tweets is attainable with excellent
performance by using state‑of‑the‑art transformer models carefully
combined with special preprocessing and well‑engineered evaluation
strategies. In cross-lingual experiments, the combination of
language-specific models (such as AraBERTv02 for Arabic) with powerful
multilingual backbones (such as XLM‑R and mBERT) enabled balanced
classification performance, and ensemble techniques offered additional
gains in stability and generalization. The decision to track
macro‑average F1 ensured fair performance measurement across biased
sentiment classes and across languages with imbalanced data
distributions.

Despite these advantages, the work was severely limited by numerous
factors. Model performance was extremely sensitive to the quality of
input, and poor‑fidelity machine translations resulted in feature
sparsity and single‑class bias in the predictions. This effect was even
more significant in low‑resource languages and in cross‑language
training scenarios where translation artifacts stacked on top.
Additionally, domain expertise and cultural sentiment cue differences
limited the ability of the models to acquire knowledge effectively
across all languages. Computational constraints also constrained
searching over wider parameter spaces and longer training times,
possibly constraining achievable accuracy.

The subsequent applications stemming from such research are manifold.
Improved translation pipelines, perhaps incorporating adaptive,
sentiment-aware MT models, can be used to decrease noise and enhance
classifier learnability. A transition into semi-supervised and
self-supervised learning methods could aid in tapping into unlabeled
multilingual data, particularly in low-resource environments. Exploring
more nuanced multimodal signals (e.g., integrating tweets with emojis,
metadata, or linked media) could pick up nuanced sentiment signals not
present in text. Lastly, fine-grained domain adaptation techniques and
cross-lingual continual learning can additionally enhance
generalization, allowing for models that continue to be accurate under
changing language usage and fluctuating norms of sentiment.

\bibliographystyle{IEEEtran}
\bibliography{references}

\end{document}